%% file: main.tex
\documentclass{article}

\usepackage{arxiv}

\usepackage{cite}
\usepackage{amsmath,amssymb,amsfonts}
\usepackage{algorithmic}
\usepackage{graphicx}
\usepackage{textcomp}
\usepackage{xcolor}
\usepackage{url}
\usepackage{float}
\usepackage{subfigure}

\newcommand{\eat}[1]{}

\begin{document}

\title{Optimizing Vessel Trajectory Compression}

\author{
 Giannis Fikioris \\
  Institute of Informatics \& Telecommunications, NCSR Demokritos\\
  Athens, Greece\\
  \texttt{gfikioris@iit.demokritos.gr} \\
  \And
 Kostas Patroumpas \\
  Information Management Systems Institute, Athena Research Center\\
  Athens, Greece\\
  \texttt{kpatro@athenarc.gr} \\
  \And
 Alexander Artikis \\
  Department of Maritime Studies, University of Piraeus\\
  Institute of Informatics \& Telecommunications, NCSR Demokritos\\
  Athens, Greece\\
  \texttt{a.artikis@unipi.gr} \\
}

\maketitle

\begin{abstract}
In previous work we introduced a trajectory detection module that can provide summarized representations of vessel trajectories by consuming AIS positional messages online. This methodology can provide reliable trajectory synopses with little deviations from the original course by discarding at least 70\% of the raw data as redundant. However, such trajectory compression is very sensitive to parametrization.
In this paper, our goal is to fine-tune the selection of these parameter values. We take into account the type of each vessel in order to provide a suitable configuration that can yield improved trajectory synopses, both in terms of approximation error and compression ratio. Furthermore, we employ a genetic algorithm converging to a suitable configuration per vessel type. Our tests against a publicly available AIS dataset have shown that compression efficiency is comparable or even better than the one with default parametrization without resorting to a laborious data inspection.
\end{abstract}

\input{introduction}

\input{related}

\input{compression}

\input{optimization}

\input{results}

\input{future}

\section*{Acknowledgments}

This work has received funding from the EU Horizon 2020 RIA program INFORE under grant agreement No 825070.

\bibliographystyle{IEEEtranS} 
\bibliography{main}

\end{document}

%% file: introduction.tex
\section{Introduction}

Thanks to the Automatic Identification System (AIS), tracking vessels across the seas provides a powerful means for maritime safety and environmental protection. However, large amounts of streaming AIS positional updates from vessels can hardly contribute additional knowledge about their actual motion patterns. Vessels are generally expected to maintain straight, predictable routes at open sea, except in cases of adverse weather conditions, accidents, traffic restrictions, etc.
In \cite{[PAA+17]} a maritime surveillance system was introduced, involving a \textit{trajectory detection} module that can provide summarized representations of vessel trajectories by consuming AIS positional messages {\em online}. The key idea behind the proposed summarization is that keeping only some {\em critical points} may be enough to reconstruct with tolerable accuracy the original course of each vessel. Indeed, instead of retaining every incoming position for every vessel or even applying a costly multi-pass trajectory simplification algorithm, this method drops positions along trajectory segments of ``normal'' motion characteristics. In addition, the retained critical points can be marked with suitable {\em annotations}, i.e., indicating stops, turning points, changes in speed, etc. The resulting {\em trajectory synopsis} per vessel is derived from those judiciously annotated critical points and can approximately reconstruct its original course. This framework has been further enhanced to run against scalable streaming data in cluster infrastructure and enable Big Data Analytics \cite{[VVS+18]} not only in the maritime domain, but also in aviation \cite{[PPT18]}. 

This methodology can provide reliable trajectory synopses with little deviations from the original course by discarding at least 70\% of the raw data as redundant;
and in case of frequent updates, even less than 1\% of the data need be retained. However, such trajectory compression is very sensitive to {\em parametrization}. Multiple rules and spatiotemporal conditions are examined for each incoming AIS location per vessel, e.g., difference in heading, speed, acceleration, etc. with respect to its previous location or its known motion pattern. So, ill selection of parameter values may have a strong impact on the quality of the resulting trajectory synopses, the achieved compression ratio, or both. Default parameter values for extracting synopses can be assigned after painstaking exploration of a given dataset with the valuable advice of domain experts. Still, in any other AIS dataset, this process may need be repeated. Last, but not least, this kind of trajectory summarization applies the same parametrized conditions over all monitored vessels, irrespective of their type, tonnage, length, etc. This ``one size fits all" approach lacks flexibility and cannot easily cope with the varying mobility patterns of vessels at sea.

In this paper, our goal is to fine-tune the selection of these parameter values. We take into account the type of each vessel (passenger, cargo, fishing, etc.) in order to provide a suitable configuration that can yield improved trajectory synopses, both in terms of approximation error and compression ratio. As many parameters are involved in this optimization problem, we employ a genetic algorithm that iterates over several combinations of the parameter values until converging to a suitable configuration per vessel type. This offline fine-tuning of parameters can then be used by the online trajectory summarization method with significant gains in terms of approximation quality and space savings. Overall, our tests against a publicly available AIS dataset~\cite{[RDC+19]} have shown that compression efficiency is comparable or even better than the one with default parametrization without resorting to a laborious data inspection.

The remainder of this paper proceeds as follows. Section~\ref{sec:related} surveys related work. Section~\ref{sec:compression} provides an overview of the mobility events employed in online summarization of evolving trajectories of vessels. Section~\ref{sec:optimization} analyzes the suggested methodology for fine-tuning the parameters used in trajectory compression, while Section~\ref{sec:customization} discusses the customization of the genetic algorithm. Section~\ref{sec:experiments} reports results from our empirical study against a real AIS dataset. Finally, Section~\ref{sec:future} summarizes the paper and outlines future research directions.

%% file: related.tex
\section{Related Work}
\label{sec:related}

Our approach on trajectory synopses over streaming AIS positions involves a kind of {\em online} path simplification. Offline techniques like \cite{[LWJ14],[MOH+14]} cannot apply, since complete trajectories must be available in advance. In contrast, we consider evolving trajectories where fresh locations are received online. Samples retained in the synopses should keep each compressed trajectory as much closer to the original one, as in fitting techniques \cite{[CWT06],[MB04]}, which  minimize approximation error. The one-pass approach in \cite{[MB04]} discards points buffered in a sliding window until the error exceeds a given threshold. The notion of safe areas in \cite{[PPS06]} keeps samples that deviate from predefined error bounds regarding speed and direction. 
Dead-reckoning policies like \cite{[WSCY99]} and mobility tracking protocols in \cite{[LDR11]} are usually employed on board of the moving objects, hence they do not seem applicable for AIS position reports. More recently, the focus on online trajectory simplification is focused on error bounds. For example, the ageing-aware approach in~\cite{[LZS+16]} uses a bounded quadrant mechanism to chose samples for the summary. Local distance checks and optimizations for higher compression are employed in~\cite{[LMZ+17]}. A novel spatiotemporal cone intersection technique involving Synchronous Euclidean Distance (SED) is applied in \cite{[LJM+19]} achieving more aggressive compression. A survey and empirical study of various trajectory simplification techniques is available in \cite{[ZDY+18]}.

Even though such generic online methods can provide reliable summaries over vessel trajectories, they entirely lack support for mobility-annotated features in the retained samples. The
maritime surveillance platform in~\cite{[PAA+17]} aims at tracking vessel trajectories and also recognizing complex events (e.g., suspicious vessel activity). Its trajectory compression  module applies a sliding window over the streaming positions, periodically reports annotated ``critical'' points (stop, turn, speed change, etc.) that need be retained in each vessel's synopsis. Tests indicated that less than 5\% of the raw data suffice to offer reliable trajectory approximations. This framework has been further enhanced as a stream-based application, coined {\em Synopses Generator}, to detect mobility events with richer semantics (multiple annotations per location, more refined conditions) with minimal latency in modern cluster infrastructures
\cite{[VVS+18]}. With extra rules, the reported mobility events can also act as notifications to timely trigger detection of more {\em complex events}~\cite{[PAD+19]}, analogous to those applied by CEP-traj \cite{[TVB+15]}. However, this framework requires careful parametrization of its various conditions, which we aim to fine tune in this paper using a genetic algorithm.

%% file: compression.tex
\section{Online Summarization of Vessel Trajectories}
\label{sec:compression}

In this Section, we overview the mobility events detected along vessel trajectories and the parameters involved in their specification by the Synopses Generator. In depth analysis of the applied trajectory summarization is available in~\cite{[PAA+17],[PPT18]}.

The Synopses Generator accepts AIS positional messages (of types 1, 2, 3, 18, 19, 27) and extracts the following attributes: \textit{MMSI}, \textit{Longitude}, \textit{Latitude}, and \textit{Timestamp}. However, this data is not error-free \cite{[INR15]}. To reduce this {\em noise}, a series of single-pass heuristics is applied to filter incoming locations online and can actually eliminate up to $20$\% of the raw AIS positions~\cite{[PAA+17]}. Even though some locations could be misreported as noisy, this hardly affects the resulting trajectory synopses as their critical points can still be identified from the fresh locations that keep continuously arriving.

Noiseless positional updates per vessel are chronologically buffered in memory. In particular, the {\em most recent portion} of each evolving trajectory is available as a distinct sequence of $m$  `clean', time-ordered locations (e.g., $m=5$). From this batch of locations, the {\em mean velocity} $\overrightarrow{v}_{m}$ of that vessel can be estimated, as well as several derived spatiotemporal features (distance, travel time, overall change in heading, etc.). To avoid considering obsolete locations in velocity calculations, a maximum {\em historical timespan} $\omega$ is set, e.g., discarding any locations received one hour before current timestamp $\tau_{now}$.

The Synopses Generator applies single-pass heuristics for detecting mobility events of the following types:

\begin{itemize}
\item {\em Stop} indicates that a vessel remains stationary by checking whether its instantaneous speed $v_{now}$ is below a threshold $v_{min}$ (e.g., 0.5 knots) over a period of time. The first location qualifying to this condition is annotated as $\mathit{stopStart}$, while the last one is marked as $\mathit{stopEnd}$. To cope with the side-effects of sea drift, vessel agility or GPS discrepancies, any fresh locations are ignored while no significant displacement is observed (less than a {\em distance} threshold $D$, e.g., 50 meters) from the previous one. If the next location is found more than $D$ meters away, the stop event ends. Overall, two critical points are issued in order to capture the duration of this phenomenon; any intermediate locations are discarded from the synopsis.

\item {\em Slow motion} means that during some time interval a vessel consistently sails at low speed below a threshold $v_{\theta}$ (e.g., $<5$ knots). The first and last AIS positions in such subtrajectory should be annotated as critical points to indicate the duration of this event.

\item {\em Change in Heading}: Once there is a difference in heading of more than a given angle threshold $\Delta\theta$ (e.g., $>4^o$) with respect to mean velocity $\overrightarrow{v}_{m}$, the previously reported location marks a turning point and qualifies as critical. Since vessels usually make smooth turns due to their large size and safety regulations, a series of such turning points may be issued to better approximate this course.

\item {\em Speed change} indicates that the vessel has either accelerated or decelerated. Such critical points are detected once the rate of change for speed deviates more than a given threshold $\alpha$ (e.g., 25\%) from its mean speed $v_{\text{m}}$ computed along the past $m$ locations, i.e. $\left| \frac{v_{\text{now}}-v_{\text{m}}}{v_{\text{now}}} \right| > \alpha$. To mark the duration of this event, one critical point ($\mathit{speedChangeStart}$) is issued when a vessel starts speeding up or slowing down; once the speed stabilizes, the respective location is marked as $\mathit{speedChangeEnd}$.

\item {\em Communication gaps} are detected when a vessel has not relayed any fresh AIS position over a time period $\Delta T$, e.g., the past 10 minutes. A pair of locations are marked as critical: $\mathit{gapStart}$ to signify when contact was lost, and $\mathit{gapEnd}$ once it gets restored.

\end{itemize}

\begin{figure}[t]
\centering
{\fbox{\includegraphics[width=85mm]{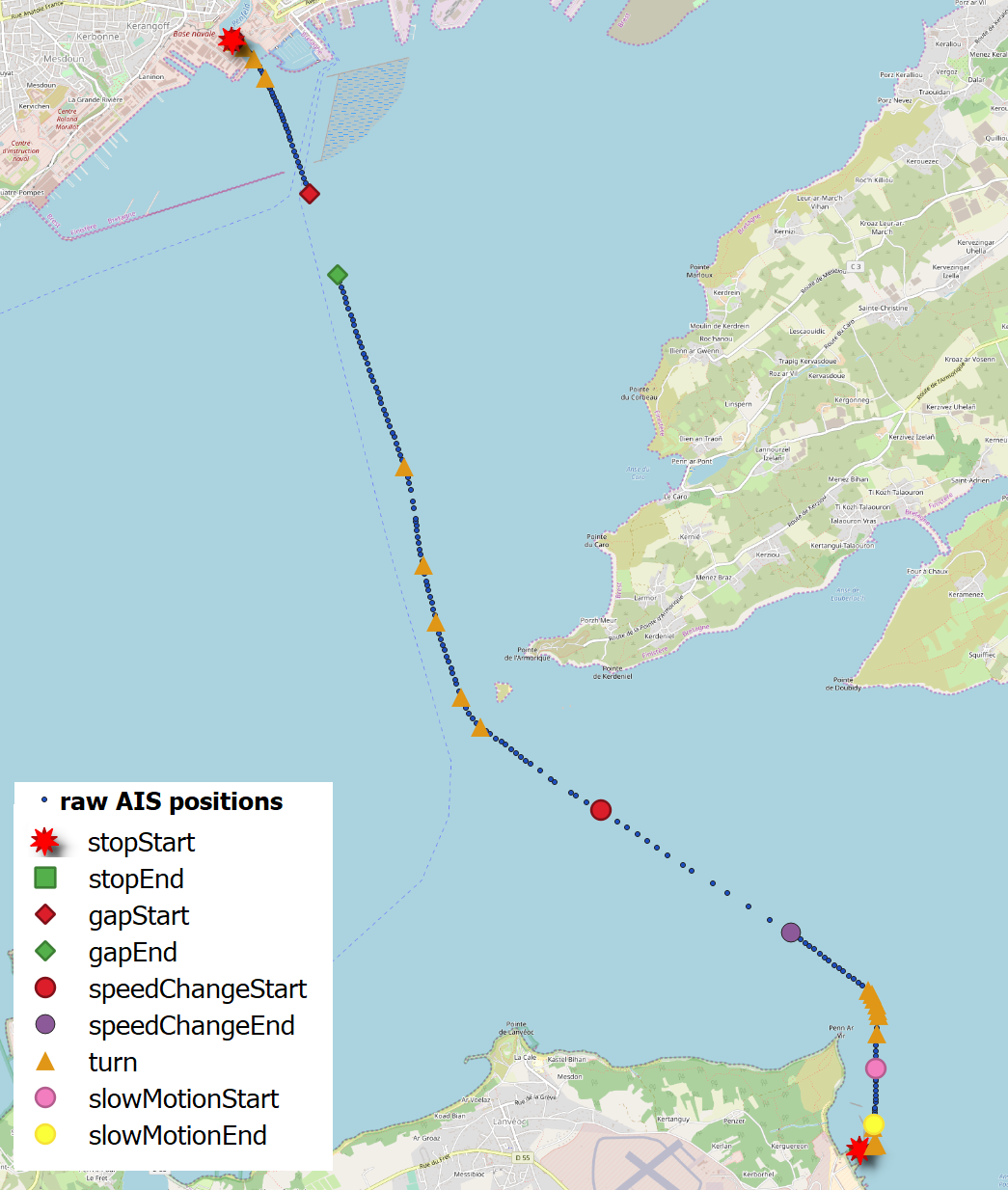}}}
\caption{Critical points detected along a vessel trajectory.} \label{fig:trajectories}
\end{figure}

Figure \ref{fig:trajectories} illustrates the process of critical point detection. 
Note that such critical points are issued at operational latency, i.e., in milliseconds. This derived stream of trajectory synopses keeps in pace with the incoming raw streaming data, which gets incrementally annotated with semantically important mobility events. This module can also achieve dramatic compression (sometimes keeping even less than 1\%) of the raw streaming data with tolerable error in the resulting approximation. 
The actual course of a vessel can be approximated using time-based interpolation to estimate positions not retained as critical \cite{[PAA+17]}.

%% file: optimization.tex
\section{Fine-tuning of Compression Parameters}
\label{sec:optimization}

Trajectory compression is very sensitive to {\em parametrization}. Table \ref{tab:parameters} presents the parameters and their default values, which have been set with the valuable advice of domain experts specifically for an AIS dataset in Brest area~\cite{[RDC+19]}. However, in any other AIS dataset, different values may be needed, depending on the geographic area, the sampling frequency, the types of monitored vessels, etc. Moreover, this kind of trajectory summarization applies the same parametrized conditions over all vessels in the data, irrespective of their type, tonnage, length, etc. This  approach lacks flexibility and cannot cope with the varying mobility patterns of the vessels. For example, a larger ship takes turns more smoothly, while a fishing or tug boat can make sharper turns. This would allow more relaxed (i.e., greater) angle threshold for larger ships without harming the quality of their synopses. On the other hand, having a stricter angle  threshold for smaller ships would entail a more accurate approximate route.

In this paper, we address such issues by taking into account the type of each vessel (passenger, cargo, fishing, etc.) in order to provide a suitable configuration that can yield improved trajectory synopses. We employ a genetic algorithm that iterates over several combinations of the parameter values until converging to a suitable configuration per vessel type. This offline fine-tuning of parameters can then be used by the online trajectory summarization method with significant gains in terms of approximation quality and space savings. 

\begin{table}[!t]
\centering
\caption{Parameters involved in synopses of vessel trajectories}\label{tab:parameters}
\begin{tabular}{clcc}
\hline
{\em Symbol} & {\em Parameter}           & {\em Value range} & {\em Default value} \\ \hline
$\Delta\theta$ & Angle threshold ($^o$)  & $2\dots25$ & 4 \\
$m$  & Buffer size (locations)  & $3\dots50$   & 5      \\
$\Delta T$ & Gap Period  (seconds) & $200\dots5000$ & 1800 \\
$\omega$ & Historical timespan  (seconds) & $300\dots5000$  & 3600     \\
$v_{min}$ & No speed threshold  (knots) & $0.05\dots2$   & 0.5 \\
$v_{\theta}$  & Low speed threshold (knots) & $0.05\dots8$  & 5 \\
$\alpha$ & Speed ratio  & $0.01\dots0.8$   & 0.25  \\
$D$ & Distance threshold  (meters) & $2\dots100$   & 50     \\
\hline
\end{tabular}
\end{table}

\subsection{Optimization Function}

We want to keep as few AIS positions as possible, i.e., minimize the \emph{compression ratio}, while at the same time we also need to minimize the \emph{approximation error} in the resulting trajectory synopses. 
The compression ratio is defined as:
\begin{equation*}
\textit{Ratio} = \frac{\textit{number of critical points for all vessels}}{\textit{number of noiseless points for all vessels}}
\end{equation*}
The parameters that we optimize do not affect the noise reduction filters; thus, the number of noiseless positions is fixed for a given input dataset of AIS messages.

Typically, the approximation error is quantified with the Root Mean Square Error (in meters):
\begin{equation*}
      \textit{RMSE} = 
      \sqrt{\frac{\sum_{p\in\textit{noiseless points}}   H^2(p, p')}
      {\textit{number of noiseless points for all vessels}}}
\end{equation*}
where $H(\cdot)$ is the Haversine distance between each original location $p$ from its time-synchronized point $p'$ in the synopsis.

We define our optimization function as
\begin{equation}\label{eq:optfunc}
   \left(\textit{RMSE + r}\right)^n \times \textit{Ratio}
\end{equation}
where $n$ and $r$ are a hyper-parameters. A term $r$ has to be added to \emph{RMSE} in order to avoid solutions to the optimization problem in which \emph{RMSE} becomes 0 by opting for no compression at all. 
No such term is needed for \emph{Ratio} because the ranges of compression parameters prevent \emph{Ratio} to reach $0$.

\begin{figure}[!t]
   \centerline{\includegraphics[width=0.5\columnwidth]{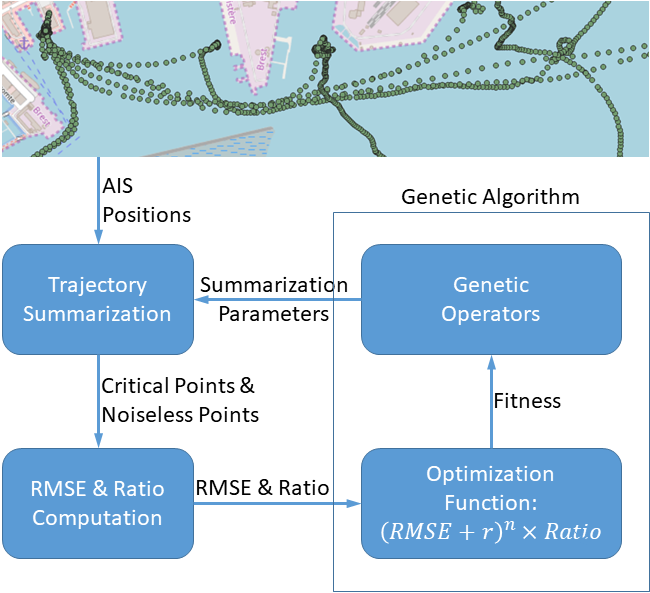}}
   \caption{Processing flow to fine-tune parameters for trajectory compression.}
   \label{fig:algorithm}
\end{figure}

\subsection{Optimization Algorithm}

Figure~\ref{fig:algorithm} illustrates our approach using a genetic algorithm to optimize the trajectory compression parameters per vessel type (Table~\ref{tab:parameters}). Genetic algorithms can solve optimization problems that are often intractable because of their large space of parameters. 
In each generation, the algorithm keeps a population of \textit{individuals}. In our case, each individual is a tuple that contains specific values for the trajectory compression parameters. The individuals of the first generation are picked using a uniform distribution. Then, the following genetic operators are applied, until convergence: selection, crossover and mutation. In the next section, we describe the customization of this genetic algorithm over an AIS dataset.

%% file: results.tex
\section{Customization of the Genetic Algorithm}
\label{sec:customization}

We trained the genetic algorithm on the publicly available AIS dataset of Brest, France \cite{[RDC+19]}. We divided these locations by vessel type and evaluated our approach on the types with the most AIS messages, as listed in Table~\ref{tab:vessel_info}.

For each vessel type, we performed 6-fold cross-validation, where each of the 6 parts of the dataset contains the same number of raw AIS locations. We compare the {\em RMSE} and the compression {\em Ratio} derived from the genetic algorithm with those achieved by the default values of the trajectory compression parameters. Recall that the default values (Table~\ref{tab:parameters}) are applied to all vessel types, while the genetic algorithm is trained separately for each vessel type.

The genetic algorithm was customized as follows:

\begin{itemize}
   \item We used \textit{Tournament Selection} of size 3, which repeatedly and randomly picks 3 individuals and selects the one with the best fitness, until the desired number of total individuals has been picked.
   \item We adopted \textit{single-point crossover}, with a crossover probability of $0.4$.
   \item We employed \textit{Gaussian Mutation}, which adds random Gaussian noise to each value of an individual with some probability. The mutation probability was set to $0.8$ since the Gaussian Mutation restricts mutation. The probability that each trait of an individual would be mutated was set to $0.5$.
\end{itemize}

In a preliminary stage to set hyper-parameters $r$ and $n$ of our optimization function in \eqref{eq:optfunc}, we selected a small part of the dataset and trained the genetic algorithm for different combinations of $r$ and $n$. We chose values for these hyper-parameters so that {\em RMSE} and compression {\em Ratio} fall below some specified thresholds. Table \ref{tab:vessel_info} presents the thresholds used for each vessel type. As shown in this table, we used three different combinations of thresholds. Each combination was chosen based on the movement and size of each vessel type in order to ensure that the derived trajectory approximation is both reliable and lightweight. For example, fishing boats often move in a zig-zag manner, which requires a large number of critical points to be annotated, while passenger boats are quite large, making larger RMSE values more tolerable. Table \ref{tab:vessel_info} presents the values of $r$ and $n$ for each vessel type as discovered by this preprocessing step.

\section{Empirical Analysis}
\label{sec:experiments}

\begin{table*}[!t]
\centering
\caption{Dataset and Genetic algorithm configuration per vessel type}
\resizebox{1.0\columnwidth}{!}{
\begin{tabular}{l|cc|cc|cc|c}
                  &\multicolumn{2}{c|}{{\em Brest dataset information}}                    & \multicolumn{2}{c|}{{\em Threshold}} & \multicolumn{2}{c|}{{\em Hyper-parameters}} & {\em Training Cost} \\
\hline
{\em Vessel type} & {\em AIS messages} & {\em Vessel count} & {\em RMSE} & {\em Ratio} & $r$ & $n$ & {\em Mean time $\pm$ Standard deviation}\\
\hline
Passenger         & 4,792,487          & 17                 & 30m        & 10\%        & 17  & 0.8 & 5.2 hours $\pm$ 16 minutes \\
Unknown      & 3,466,765          & 115                & 15m        & 15\%        & 10  & 1.0 & 4.5 hours $\pm$ ~8 minutes \\
Fishing           & 3,288,577          & 161                & 30m        & 30\%        & 17  & 0.7 & 4.5 hours $\pm$ 39 minutes \\
Tug               & 1,411,761          & 15                 & 15m        & 15\%        & ~2  & 1.6 & 1.8 hours $\pm$ ~4 minutes \\
Cargo             & 1,198,228          & 184                & 30m        & 10\%        & 13  & 0.8 & 1.5 hours $\pm$ ~2 minutes \\
Military          & ~~802,045          & 12                 & 15m        & 15\%        & 10  & 1.4 & ~~1 hour~ $\pm$ ~3 minutes \\
\hline
\end{tabular}} \label{tab:vessel_info}
\end{table*}

Next, we report results from an empirical study of the proposed methodology using a genetic algorithm (GA) towards fine-tuning of parameters for trajectory compression.

\subsection{Experimental Setup}
\label{sec:setup}

The Synopses Generator\footnote{ \url{https://github.com/DataStories-UniPi/Trajectory-Synopses-Generator}} is developed in Scala using the DataStream API of {\em Apache Flink} with {\em Apache Kafka} as broker for streaming messages. The genetic algorithm for fine-tuning the parameters of Synopses Generator is implemented in Python, using the evolutionary framework DEAP \cite{Deap}.
All tests were conducted on an Intel\textregistered{} Core\texttrademark{} i7-4790 CPU @ 3.60GHz machine with 16GB RAM.

\subsection{Experimental Results}

Regarding the training cost of GA, we measured the time required to train five sixths of the dataset, for each run of the 6-fold cross-validation. Table \ref{tab:vessel_info} reports the average training time and the standard deviation of the cost among runs. As expected, there is a direct correspondence between dataset size (in terms of AIS messages per vessel type) and this cost.

\begin{figure*}[!t] 
\centering
\subfigure[Passenger ships]{
    \includegraphics[width=0.32\textwidth]{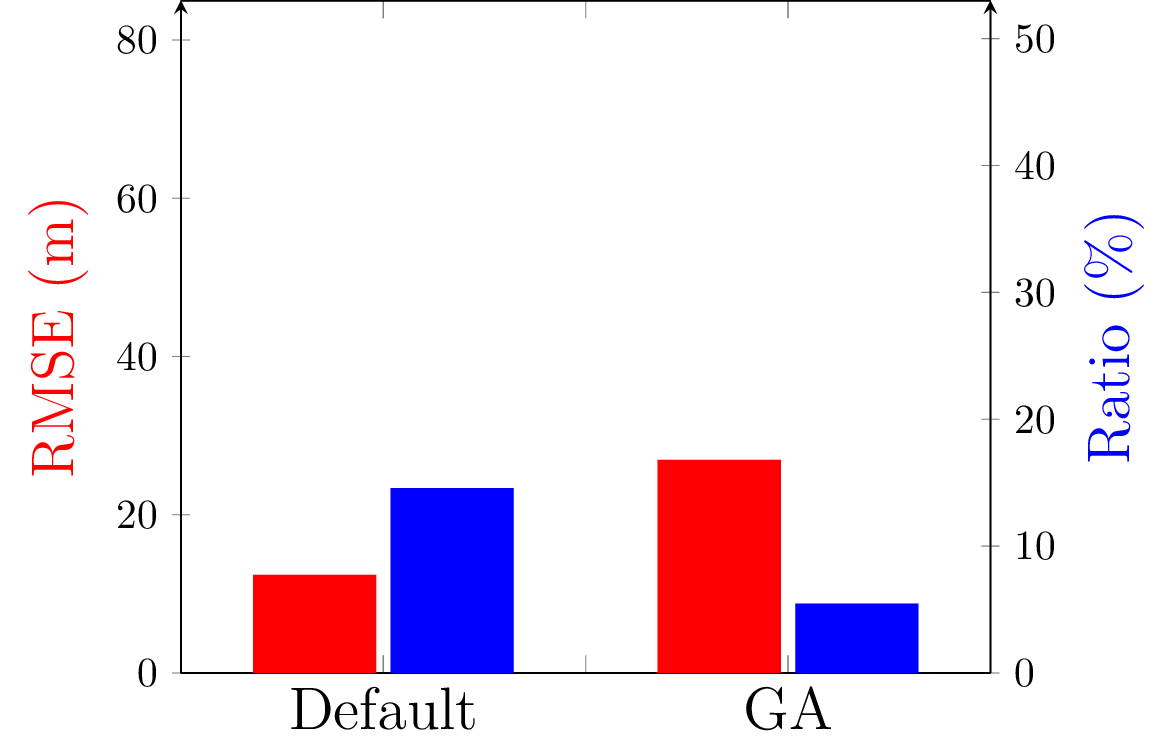}
}\label{fig:passenger}%
\subfigure[Unknown type]{
    \includegraphics[width=0.32\textwidth]{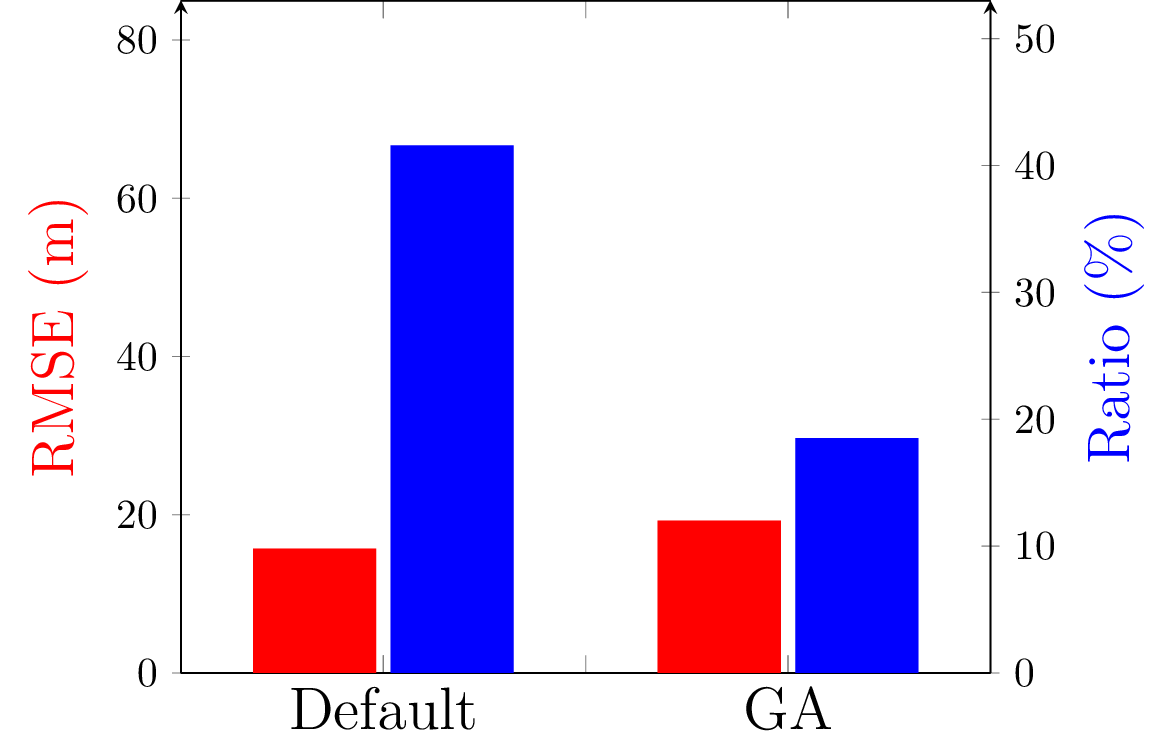}
}\label{fig:unidentified}%
\subfigure[Fishing vessels]{
    \includegraphics[width=0.32\textwidth]{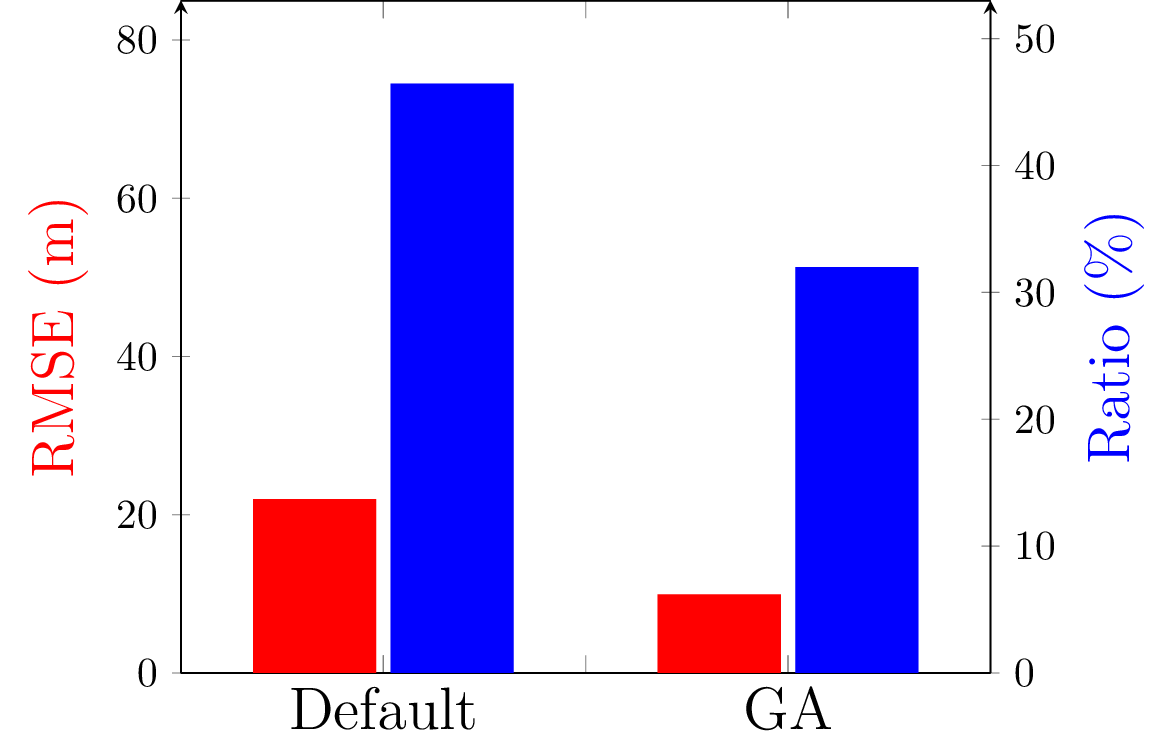}
}\label{fig:fishing}%

\subfigure[Tug boats]{
    \includegraphics[width=0.32\textwidth]{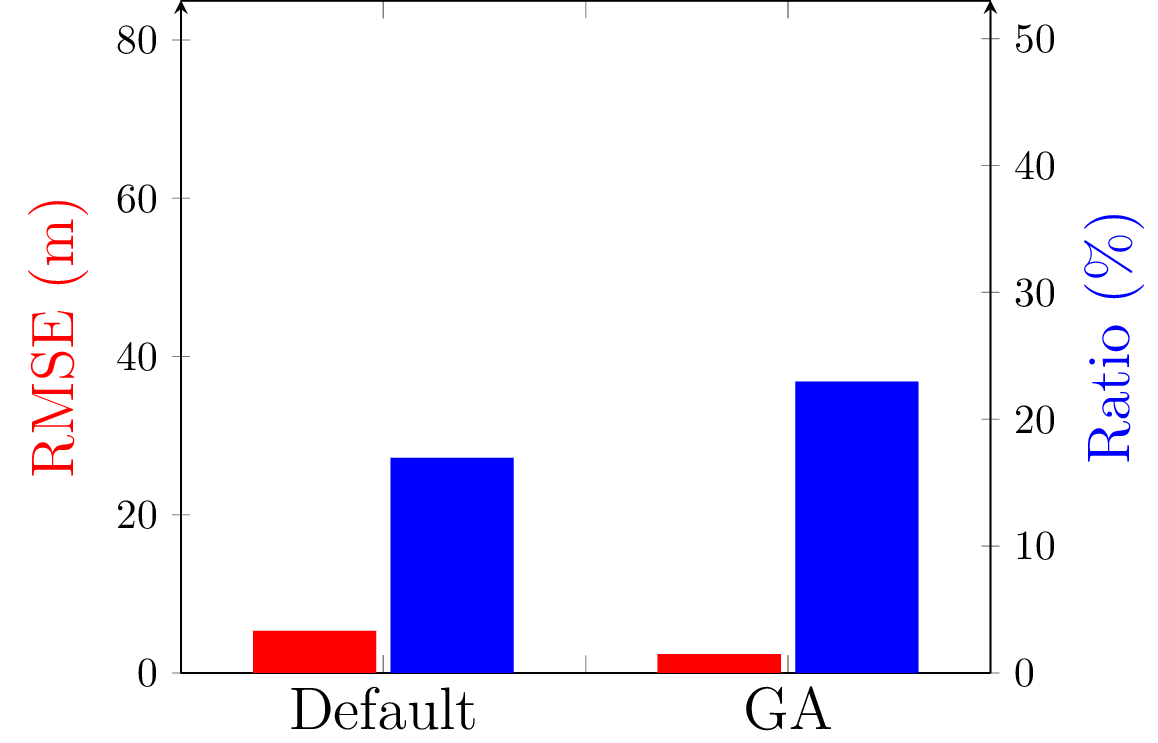}
}\label{fig:tug}%
\subfigure[Cargo ships]{
    \includegraphics[width=0.32\textwidth]{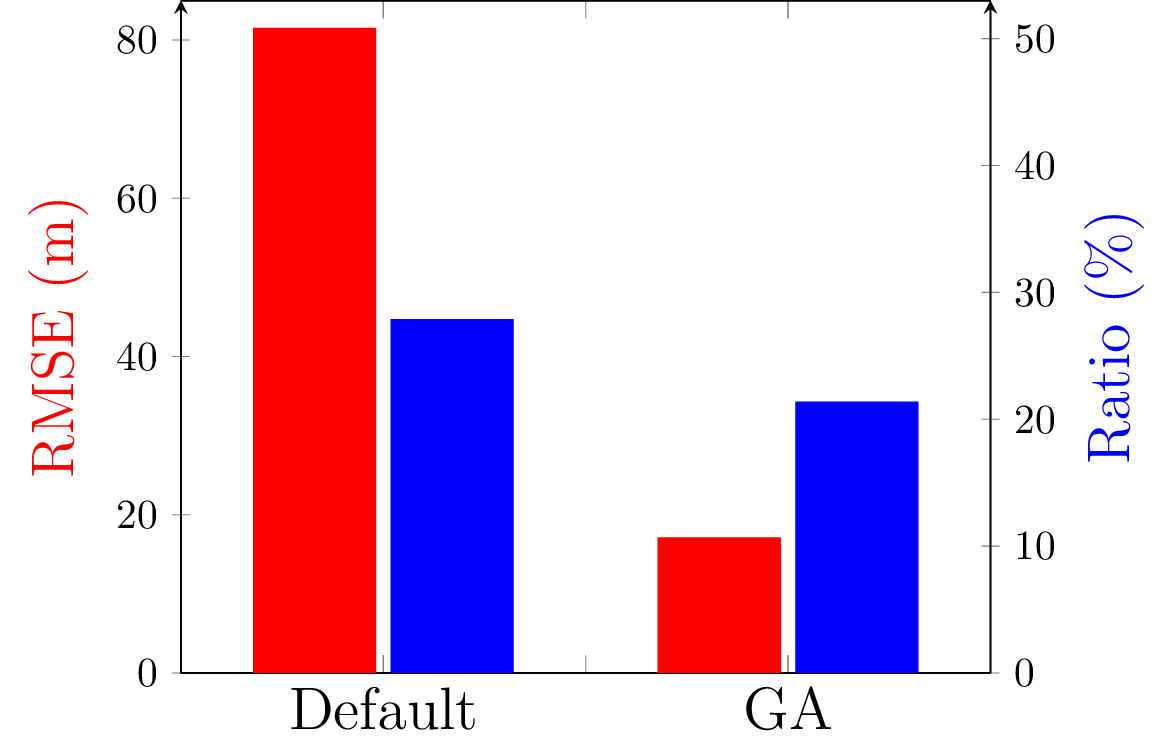}
}\label{fig:cargo}%
\subfigure[Military vessels]{
    \includegraphics[width=0.32\textwidth]{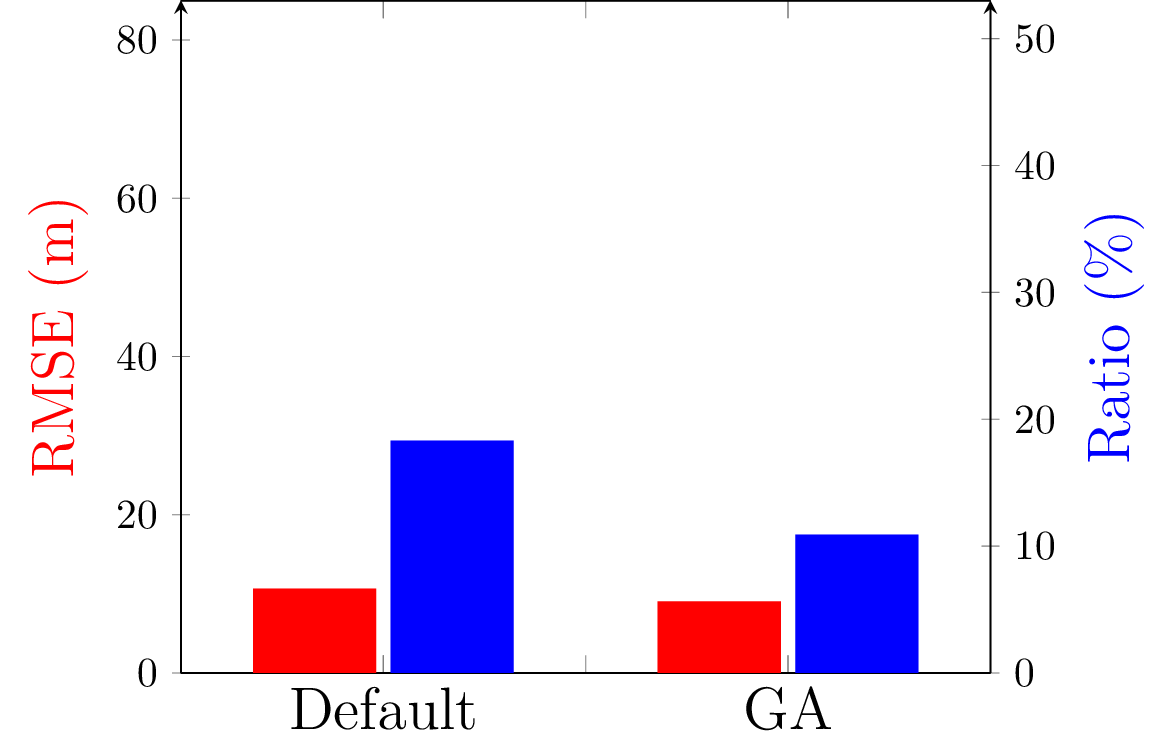}
}\label{fig:military}%
\caption{Comparison of compression efficiency for the main vessel types available in the Brest dataset. In each plot, the two bars on the left concern the use of the default parameter values, while the two bars on the right represent the results from the genetic algorithm (GA). }
\label{fig:results}
\end{figure*}

With respect to compression efficiency, Figure~\ref{fig:results} compares results for {\em RMSE} (red bars) and compression {\em Ratio} (blue bars) between the default parametrization and the fine-tuned one suggested by the GA.
Table~\ref{tab:finalparams} lists the fine-tuned parameter values for each vessel type, i.e., those that achieved the lowest score according to \eqref{eq:optfunc} over all folds.

\subsubsection{Passenger ships} 
These generally follow straight courses at open seas, making their synopses easier to maintain. The default parameters achieve a rather low RMSE of about $15$m. As passenger ships often generally have quite large sizes, both in width and in length, such low RMSE values are rather strict and lead to poorer compression. GA takes advantage of this, by only increasing RMSE to an acceptable value just under $30$m and manages to keep only a third of the critical points in the synopses. This improvement is possible according to the thresholds specified in Table~\ref{tab:vessel_info}.

Regarding the fine-tuned parameters (Table~\ref{tab:finalparams}), note that less effort is made to keep critical points. The values for \textit{angle threshold} ($\Delta\theta$) and \textit{speed ratio} ($\alpha$) are much higher than the default ones, leading to less detailed, yet fairly accurate synopses compared to the ones created with the default values.

\subsubsection{Vessels of unknown type}
Default parametrization yields a fair RMSE at the expense of a very poor compression, which keeps more than 60\% of the original AIS locations. In contrast, GA provides a comparable RMSE and reduces by half the number of critical points. This effect should be attributed to the unsuitability of default values for this subset of the data. From Table~\ref{tab:finalparams}, it is clear that GA relaxes the \textit{angle threshold ($\Delta\theta$)} and suggests a lower value for \textit{speed ratio ($\alpha$)}. Instead, under the default parametrization (Table~\ref{tab:parameters}), it seems that the resulting synopses unnecessarily kept too many critical (especially {\em turning}) points, whereas several {\em change-in-speed} points were missed.

\subsubsection{Fishing vessels}
In general, such trajectories are far from straight, thus a larger fraction of original points must be kept as critical  to sustain a reasonable approximation error. With the default parameters, both RMSE and compression ratio are high compared to other vessel types. Instead, the optimal parameter values discovered by the GA offer much better synopses: Ratio is close to the desired threshold of $30\%$, while RMSE falls substantially under the threshold value of $30$ meters.

This considerable improvement is mostly due to suitable selection of smaller values for \textit{buffer size} ($m$) and \textit{distance threshold} ($D$) in Table~\ref{tab:finalparams}, compared to the default ones (Table~\ref{tab:parameters}). Since fishing boats often make zig-zag or circular manoeuvres, larger values in these parameters (as in the default configuration) may lead to inaccurate estimates of their mean velocity, thus yielding many extra critical points.

\subsubsection{Tug boats}
Tug boats move mainly inside the port at low speed, making it easier to create an accurate synopsis with few points. This is also noticeable in the very small RMSE and low compression ratio achieved with the default values. GA achieves an even better RMSE with only a small increase of the compression ratio. Note that the ratio value slightly exceeds the threshold of $15\%$, most likely because the dataset used for hyper-parameter tuning might not be enough representative of the motion patterns overall.

Fine-tuning suggests very small values for parameters \textit{angle threshold} ($\Delta\theta$) and \textit{speed ratio} ($\alpha$) in Table~\ref{tab:finalparams}, probably due to the simple motion paths observed for this type of vessels. This might seem surprising given that the ratio threshold was set to  $20\%$ only, but it actually makes sense: tug boats move slowly with few changes in heading and speed, thus rarely generating critical points, even with low parameter values.

\subsubsection{Cargo ships}
The pattern of movement for such vessels is different from other types. This is confirmed by the unusually high RMSE and moderate compression ratio achieved with default parametrization. GA improves a little the compression ratio, but manages to reduce RMSE by a large margin. 

These results actually validate our main argument: different mobility patterns require different degrees of compression. Admittedly, default values may offer fair compression for several vessel types, but not for the larger and bulkier cargo vessels. GA is able to ``learn'' this different behaviour, and subsequently chooses suitable values for the parameters in order to offer more reliable synopses. Note in Table~\ref{tab:finalparams} that the \textit{distance threshold} ($D$) takes a very low value. Like in most other vessel types, the \textit{speed ratio} ($\alpha$) value is as low as possible. Instead, under default parametrization, $\alpha$ was relatively high and the synopses failed to keep enough \textit{change-in-speed} critical points, hence the unusually increased RMSE.

\subsubsection{Military vessels}
Just like tug boats, movement of military ships is also rather simple, as manifested by the low RMSE and ratio that the default parameters can offer. With the GA, we manage to decrease both measures by a small amount, while also well satisfying the thresholds set in Table~\ref{tab:vessel_info}. The fine-tuned parametrization (Table~\ref{tab:finalparams}) suggests a surprisingly low value for the \textit{low speed threshold} ($v_\theta$). This means that GA discovered that such vessels often move at speeds higher than $0.45$ knots, so it considers slow motion less noteworthy for the synopses. In case the value of $v_\theta$ was set higher, more critical points would be generated, scarcely offering any extra details on the approximate course, but worsening the compression ratio without any significant benefit on RMSE.

\begin{table}[!t]
\centering
\caption{Parameters picked by the genetic algorithm per vessel type}\label{tab:finalparams}
\begin{tabular}{crrrrrrr}
\hline
{\em Parameter}     & {\em Default} & {\em Passenger} & {\em Unknown} & {\em Fishing} & {\em Tug} & {\em Cargo} & {\em Military} \\ \hline
$\Delta\theta$      & 4    & 10.71 & 17.58 & 18.99 & 4.96 & 17.5  & 11.68 \\ 
$\Delta T$          & 1800 & 200   & 400   & 200   & 450  & 2500  & 2600  \\ 
$m$                 & 5    & 50    & 21    & 3     & 29   & 3     & 3     \\ 
$\omega$            & 3600 & 2750  & 1500  & 3550  & 2300 & 1750  & 4800  \\ 
$v_{\textit{min}}$  & 0.5  & 2.0   & 1.52  & 0.41  & 0.84 & 0.81  & 0.88  \\ 
$D$                 & 50   & 76.51 & 44.45 & 23.97 & 2.0  & 15.12 & 22.96 \\ 
$\alpha$            & 0.25 & 0.63  & 0.01  & 0.01  & 0.01 & 0.01  & 0.01  \\ 
$v_\theta$          & 5    & 4.58  & 1.02  & 0.61  & 6.06 & 0.82  & 0.45  \\
\hline
\end{tabular}
\end{table}

%% file: future.tex
\section{Summary and future work }
\label{sec:future}

We have been developing a trajectory detection module that can provide summarized representations of vessel trajectories by consuming AIS positional messages online.
In this paper, we presented an approach for fine-tuning the selection of the parameter values for this module. We took into account the type of each vessel in order to provide a suitable configuration that can yield improved trajectory synopses, both in terms of approximation error and compression ratio. Moreover, we employed a genetic algorithm converging to a suitable configuration per vessel type. Our empirical analysis showed that compression efficiency may be better than the one with default parametrization, without resorting to time-consuming data inspection. For future work, we intend to investigate the effects of automated parameter value selection on the performance of composite maritime event recognition \cite{[PAD+19]} based on such compressed trajectories.